\documentclass{article}
\usepackage{spconf,amsmath,graphicx,hyperref}
\usepackage{booktabs}
\usepackage{multirow}
\usepackage{xcolor}
\usepackage{adjustbox}
\usepackage{tabularx,array}
\usepackage[table]{xcolor}
\usepackage{colortbl}

\title{Confidence-Guided Error Correction for \\
Disordered Speech Recognition}

\name{
  Abner Hernandez$^{1}$,
  Tom\'as Arias-Vergara$^{1,2}$,
  Andreas Maier$^{1}$,
  Paula~Andrea P\'erez-Toro$^{1,2}$
}

\address{
  $^{1}$ Pattern Recognition Lab, Friedrich-Alexander-Universität Erlangen-Nürnberg, Germany \\
  $^{2}$ GITA Lab. Facultad de Ingeniería. Universidad de Antioquia UdeA, Medellín, Colombia
}
\begin{document}
%
\maketitle
\begin{abstract}
We investigate the use of large language models (LLMs) as post-processing modules for automatic speech recognition (ASR), focusing on their ability to perform error correction for disordered speech. In particular, we propose confidence-informed prompting, where word-level uncertainty estimates are embedded directly into LLM training to improve robustness and generalization across speakers and datasets. This approach directs the model to uncertain ASR regions and reduces overcorrection. We fine-tune a LLaMA 3.1 model and compare our approach to both transcript-only fine-tuning and post hoc confidence-based filtering. Evaluations show that our method achieves a 10\% relative WER reduction compared to naive LLM correction on the Speech Accessibility Project spontaneous speech and a 47\% reduction on TORGO, demonstrating the effectiveness of confidence-aware fine-tuning for impaired speech.
\end{abstract}
\begin{keywords}
Automatic speech recognition, error correction, large language models, confidence estimation, speech disorders
\end{keywords}
\section{Introduction}
\label{sec:intro}

Automatic speech recognition (ASR) systems have advanced rapidly, driven by deep neural architectures, large-scale transcribed corpora, and self-supervised pre-training. While they perform well across many domains and speakers, their accuracy degrades on disordered speech, such as dysarthria, which departs from typical phonetic and prosodic patterns.
Dysarthria, a motor speech disorder caused by neurological impairment, is marked by reduced articulatory precision, disrupted rhythm, abnormal prosody, and frequent pauses, prolongations, or repetitions~\cite{dysarthria,dysarthria2}. Although individuals with speech impairments could greatly benefit from ASR support, their speech is underrepresented in training data, limiting the effectiveness of current models in real-world applications.

Recent research aimed at improving ASR for dysarthric speech has primarily focused on addressing acoustic variability and limited training resources~\cite{b1,li25f_interspeech}. In contrast, LLM-based post-processing for disordered speech remains underexplored, despite their ability to use context modeling and semantic reasoning to recover meaning from degraded input. However, prior work~\cite{Hrinchuk2020, ma23e_interspeech, li2024} has shown that such models can also overcorrect when ASR outputs are already accurate, sometimes increasing word error rate (WER).

This study formulates LLM-based error correction for dysarthric speech as a sequence-to-sequence task. LLaMA models are fine-tuned on reference–hypothesis pairs from the Speech Accessibility Project (SAP) dataset using instruction prompts. ASR confidence scores are incorporated into the input to guide corrections toward uncertain regions. 

The contributions of this work are as follows:
\begin{itemize}
\item Embedding confidence scores into LLM prompts to guide selective error correction.
\item Mitigating overcorrection compared to naive LLM approaches on dysarthric speech.
\item Demonstrating generalization across datasets and ASR architectures.
\item Analyzing how confidence-guided models focus corrections on low-confidence regions while preserving high-confidence words.
\end{itemize}

\section{Related Studies}
\label{sec:related}

ASR error correction has received increasing attention with the rise of LLMs for speech applications. Early approaches trained encoder-decoder architectures to correct recognition errors~\cite{Hrinchuk2020, Guo2019}. Subsequent work investigated the use of N-best hypotheses to provide richer context during training, including the T5 model~\cite{ma23e_interspeech} and LLaMA~\cite{radhakrishnan2023}. Recent work has also integrated confidence scores into ASR error correction~\cite{pu2024,naderi24}, highlighting the importance of confidence-aware correction mechanisms.

However, the use of LLM-based error correction with disordered speech remains limited. In~\cite{yusufali}, LLMs were applied to intent prediction in text-based AAC systems for dysarthria, where adapting language model representations to individual patterns improved context sensitivity and message prediction. The SAP dataset was examined in~\cite{laquatra25_interspeech}, where a two-stage framework combined Whisper-based ASR with a Flan-T5 model for error correction. Results showed that error correction improves transcription accuracy on both structured and spontaneous speech, though challenges remain for isolated word recognition. Another study examined SAP data with instruction-tuned LLMs~\cite{hernandez2025enhancing}, where Flan-T5 and LLaMA were fine-tuned on reference–hypothesis pairs generated from Fast Conformer and Whisper outputs embedded in instructional prompts. The models achieved strong performance on structured commands but showed limited gains on spontaneous speech.

\section{Methods}
\label{sec:method}

\begin{figure}[h]
  \centering
  \includegraphics[width=0.75\linewidth]{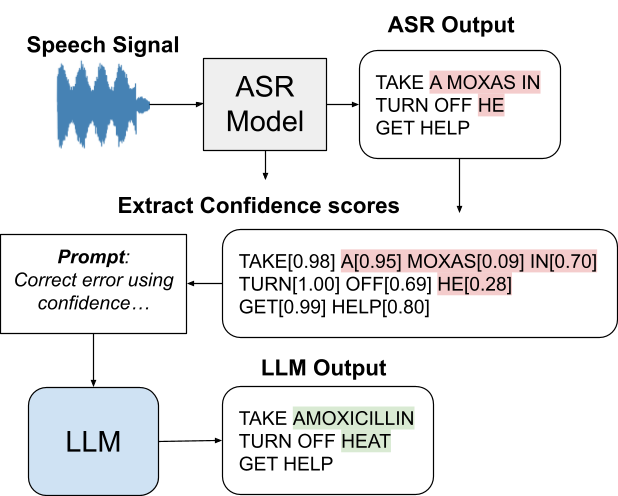}
  \caption{Pipeline for ASR error correction with confidence informed prompts.}
  \label{fig:asr-llm-conf}
\end{figure}

We propose a confidence-guided framework for ASR error correction using fine-tuned LLaMA models. Entropy-based confidence scores are computed to identify low-confidence regions in ASR transcripts, which are then used to guide LLM corrections for impaired speech, as illustrated in Fig.~\ref{fig:asr-llm-conf}.

\subsection{Impaired Speech Data}
\subsubsection{Speech Accessibility Project}
The SAP dataset used in this study is the \textit{2024-04-30 Phase 1 release} used in the \textit{Interspeech 2025 Speech Accessibility Project Challenge}~\cite{zheng25_interspeech}, containing recordings from speakers with Parkinson's disease (PD). It includes a training set of 369 speakers (around 290 hours) and a development set of 55 speakers (43.5 hours). Since the official test set is unavailable, we randomly selected a 15-speaker subset from the development set (around 9 hours) as our test set, ensuring speaker independence across training, validation, and test splits.

Development and test utterances are divided into \textit{shared} (58\%) and \textit{unshared} (42\%) categories. Shared data includes material read by all 
participants or by controlled subsets, while unshared data consists of participant-specific commands and spontaneous speech~\cite{sap}.

\subsubsection{TORGO Database}
The TORGO database~\cite{torgo} contains recordings from seven dysarthric and eight control speakers, with diagnoses including cerebral palsy and ALS. We used it solely to assess cross-database generalizability, selecting only multi-word utterances from speakers with mild to moderate dysarthria, for a total of 633 utterances.

\subsection{ASR Model Setup}
Training data are generated by running inference on the full train and development set. We use the Parakeet TDT-CTC 110M model for both generating reference–hypothesis pairs and evaluation. This lightweight model combines FastConformer~\cite{fastconformer} encoders with a hybrid Token-and-Duration Transducer (TDT)~\cite{tdt}.

To assess cross-model generalization, we also evaluate our correction models on outputs from Whisper~\cite{whisper}. Specifically, we use distil-whisper/distil-large-v3.5\footnote{\url{https://huggingface.co/distil-whisper/distil-large-v3.5}}, a distilled variant of Whisper Large v3 optimized for efficiency while retaining high transcription accuracy. Unlike Parakeet, Whisper is not used to generate training data, allowing us to evaluate whether confidence-guided correction generalizes to a different ASR architecture with distinct error patterns. It serves solely as an additional evaluation system to test cross-model transferability.

\subsection{Confidence-Based Scores}
After generating ASR hypotheses with Parakeet, we compute entropy-based confidence scores to quantify prediction reliability. Confidence estimation is performed at the frame level and then aggregated to words, enabling the correction model to identify uncertain regions in the transcript.

Entropy provides an information-theoretic measure of uncertainty based on the full token distribution~\cite{entropy}. For Gibbs entropy, the normalized confidence score $F_g(p)$ is defined as

\begin{equation}
F_g(p) = 1 - \frac{H_g(p)}{\max H_g(p)} = 1 + \frac{1}{\ln V} \sum p_v \ln(p_v),
\end{equation}

where $p$ is the predicted distribution and $V$ is the vocabulary size. While this captures general uncertainty, modern ASR models are often overconfident, motivating the use of Tsallis entropy for more flexible calibration. With entropic index $\alpha$, the Tsallis confidence score is given by

\begin{equation}
F_{ts}(p) = \frac{V^{1 - \alpha} - \sum p_v^\alpha}{V^{1 - \alpha} - 1},
\end{equation}

where smaller $\alpha$ values increase sensitivity to peaked distributions.  

Since confidence is first computed at the frame level, we aggregate these scores to obtain word-level estimates. Three aggregation strategies are considered: mean, minimum, and product. Each emphasizes different aspects of uncertainty. For example, the minimum is highly sensitive to localized uncertainty, while the mean provides a more balanced view across frames. We evaluate four correction strategies: (1) \textbf{Naive Correction} applies correction to every hypothesis, (2) \textbf{Sentence-Level Filtering} triggers correction only if overall sentence-level confidence falls below a threshold, where the sentence-level confidence is calculated as the geometric mean of all word-level confidence scores in that sentence, (3) \textbf{Word-Level Filtering} triggers correction if any word's confidence falls below a threshold, and (4) \textbf{Confidence Prompting} explicitly includes word-level confidences in the input prompt of the correction model.




\subsection{Fine-tuning LLaMA for Error Correction}
We fine-tune the 8-billion-parameter LLaMA 3.1 Instruct model on 130,000 reference–hypothesis pairs from SAP. To enable parameter-efficient adaptation, we apply low-rank adapters (LoRA), reducing the number of trainable parameters to approximately 42 million. The best performance was obtained with rank $r=16$ and scaling factor $\alpha=16$ after a grid search over $r, \alpha \in \{8,16,32\}$.

Instruction-style prompts were designed to simulate ASR correction. While the base setup asked the model to correct transcriptions directly, we extend this formulation by incorporating word-level confidence scores, instructing the model to prioritize uncertain tokens. Fig.~\ref{fig:prompt-template-conf} illustrates this confidence-informed prompt, which encourages targeted and reliable corrections while preserving correct content. All experiments were conducted on a single NVIDIA A100 GPU for 3 epochs.

\begin{figure}[h]
\centering
\footnotesize
\fbox{\begin{minipage}{0.85\linewidth}
\texttt{Instruction:} \\
You are an expert in speech-language processing and automatic speech recognition (ASR) correction, specializing in speech from individuals with dysarthria due to conditions such as PD.\\

Your task is to correct errors in the sentence below. \\
\textbf{Use the provided confidence scores to guide your decisions. Words with lower confidence are more likely to be incorrect.}

\vspace{0.3\baselineskip}

\texttt{Question:} \\
Correct the following sentence (confidence scores are shown in brackets): \\
HOW\textbf{[1.00]} MANY\textbf{[0.85]} \textbf{RAFELLES[0.61]} \\

\vspace{0.3\baselineskip}

\texttt{Correction:}
HOW MANY \textbf{REFILLS} 

\end{minipage}}
\caption{Example prompt used to train confidence-informed error correction model.}
\label{fig:prompt-template-conf}
\end{figure}

\vspace{-3mm}

\section{Results}
\label{sec:results}

We evaluate our confidence-informed ASR error correction approach on all test sets, comparing it against naive LLM correction and traditional confidence-based filtering methods. The analysis examines overall WER performance and the distribution of helpful versus harmful corrections across different confidence levels.

Tab.~\ref{tab:main-res} reports WER after applying our confidence-informed LLM correction to the outputs of each ASR system. The values in parentheses indicate the uncorrected ASR WER. 
For Parakeet outputs, correction reduces WER from 15.64\% to 4.95\% on SAP-shared (68.4\% relative) and from 9.94\% to 9.47\% on SAP-unshared (4.6\% relative). 
For Whisper outputs, correction reduces WER from 13.10\% to 4.19\% on SAP-shared (68.0\% relative) and from 18.0\% to 17.72\% on SAP-unshared (1.6\% relative).

\begin{table}[h]
\centering
\caption{WER (\%) after LLM correction using Tsallis entropy with different $\alpha$ values and aggregation methods. Parentheses in the second column show the uncorrected ASR WER.}
\scriptsize
\resizebox{\linewidth}{!}{
\begin{tabular}{
    >{\centering\arraybackslash}p{0.8cm} 
    >{\centering\arraybackslash}p{1.0cm} 
    >{\centering\arraybackslash}p{0.8cm}  
    >{\centering\arraybackslash}p{0.8cm} 
    >{\centering\arraybackslash}p{0.8cm} 
    >{\centering\arraybackslash}p{0.8cm} 
}
\toprule
\textbf{System} & \shortstack{\textbf{Test Set}\\\textbf{(ASR WER)}} & \textbf{Alpha} & \multicolumn{3}{c}{\shortstack{\textbf{LLM WER (\%)} \\ \textbf{Aggregation}}} \\
\cmidrule(lr){4-6}
 & & & \textbf{Product} & \textbf{Mean} & \textbf{Min} \\
\midrule
\multirow{12}{*}{Parakeet}
 & \multirow{4}{*}{\shortstack{SAP-shared\\(15.64\%)}} 
   & 0.9 & \textbf{4.95} & 5.21 & 5.06 \\
 &  & 0.7 & 5.11 & 5.38 & 5.34 \\
 &  & 0.5 & 5.18 & 5.39 & 5.35 \\
 &  & 0.3 & 5.07 & 5.29 & 5.16 \\
\addlinespace
 & \multirow{4}{*}{\shortstack{SAP-unshared\\(9.94\%)}} 
   & 0.9 & \textbf{9.47} & 9.56 & 9.55 \\
 &  & 0.7 & 9.51 & 9.59 & 9.52 \\
 &  & 0.5 & 9.59 & 9.57 & 9.58 \\
 &  & 0.3 & 9.48 & 9.52 & 9.54 \\
\addlinespace
 & \multirow{4}{*}{\shortstack{TORGO\\(10.83\%)}} 
   & 0.9 & 11.37 & 10.69 & 10.89 \\
 &  & 0.7 & 10.80 & 10.62 & 10.60 \\
 &  & 0.5 & 10.65 & 10.62 & \textbf{10.58} \\
 &  & 0.3 & 12.56 & 11.00 & 11.77 \\
\midrule
\multirow{12}{*}{Whisper}
 & \multirow{4}{*}{\shortstack{SAP-shared\\(13.10\%)}} 
   & 0.9 & 4.45 & 4.62 & 4.45 \\
 &  & 0.7 & 4.59 & 4.74 & 4.66 \\
 &  & 0.5 & 4.36 & 4.64 & 4.43 \\
 &  & 0.3 & \textbf{4.19} & 4.24 & 4.23 \\
\addlinespace
 & \multirow{4}{*}{\shortstack{SAP-unshared\\(18.00\%)}} 
   & 0.9 & 17.73 & 17.74 & \textbf{17.72} \\
 &  & 0.7 & 17.73 & 17.75 & 17.76 \\
 &  & 0.5 & 17.87 & 17.82 & 17.83 \\
 &  & 0.3 & 18.05 & 17.92 & 17.91 \\
\addlinespace
 & \multirow{4}{*}{\shortstack{TORGO\\(8.64\%)}} 
   & 0.9 & \textbf{8.51} & 8.66 & 8.84 \\
 &  & 0.7 & 8.95 & 8.71 & 8.79 \\
 &  & 0.5 & 11.00 & 11.05 & 11.25 \\
 &  & 0.3 & 15.40 & 14.54 & 14.47 \\
\bottomrule
\end{tabular}}
\label{tab:main-res}
\end{table}

On TORGO, applying our correction to Parakeet outputs lowers WER from 10.83\% to 10.58\% (2.3\% relative), while applying correction to Whisper outputs lowers WER from 8.64\% to 8.51\% (1.5\% relative). These results indicate that confidence-informed correction generalizes to spontaneous speech (SAP-unshared), unseen disordered-speech datasets (TORGO), and across ASR architectures.

Tab.~\ref{tab:correction-strategies} compares different error correction strategies across the three test sets with Parakeet outputs. We tested confidence thresholds ranging from 10\% to 90\% for filtering approaches, but only report the best performing thresholds. On SAP-shared, word-level filtering at a high threshold (90\%) achieved the lowest WER of 4.55\%, slightly outperforming both naive LLM correction (4.69\%) and confidence-informed prompting (4.95\%). For SAP-unshared, confidence-informed prompting obtained the best performance at 9.48\%, compared to 10.56\% with naive LLM correction. Word-level filtering also reduced errors to 9.87\% at a 50\% threshold.

On TORGO, both word and sentence-level filtering converged to a WER of 10.73\%, while confidence-informed prompting provided the strongest result at 10.58\%, a large improvement over naive LLM correction (20.01\%). Notably, naive LLM correction significantly degraded performance compared to the original ASR baseline (10.83\% to 20.01\% WER), highlighting the overcorrection problem our confidence-guided approach addresses. Overall, the results indicate that confidence-based correction is particularly useful for spontaneous speech, as in SAP-unshared, and for generalizing to datasets outside of training, such as TORGO.


\begin{table}[h]
\centering
\caption{WER Comparison of error correction strategies. Thresholds in \% are shown for word and sentence-level filtering.}
\resizebox{\columnwidth}{!}{%
\begin{tabular}{l c c c c}
\toprule
\textbf{Dataset} & \shortstack{\textbf{LLM} \\ \textbf{(Naive)}} & \shortstack{\textbf{Word-Level Filter} \\ \textbf{(Threshold)}} & \shortstack{\textbf{Sent.-Level Filter} \\ \textbf{(Threshold)}} & \shortstack{\textbf{LLM} \\ \textbf{(w/ conf.)}} \\
\midrule
SAP-shared   & 4.69  & \textbf{4.55 (90\%)}  & 8.08 (90\%)  & 4.95  \\
SAP-unshared & 10.56 & 9.87 (50\%)  & 9.94 (80\%)  & \textbf{9.48}  \\
TORGO        & 20.01 & 10.73 (60\%) & 10.73 (80\%) & \textbf{10.58} \\
\bottomrule
\end{tabular}%
}
\label{tab:correction-strategies}
\end{table}


\subsection{Analysis of Corrections}

Tab.~\ref{tab:confidence-aware-corrections} shows consistent confidence-aware behavior using dataset-specific mean confidence as thresholds. The LLM applies helpful corrections more often to low-confidence utterances: 63.8\% vs. 17.5\% (SAP-shared), 24.3\% vs. 10.1\% (SAP-unshared), and 9.6\% vs. 2.4\% (TORGO). The model shows selective intervention with higher correction attempt rates for low-confidence sentences: 74.7\% vs. 20.3\% (SAP-shared), 53.3\% vs. 18.5\% (SAP-unshared), and 30.9\% vs. 4.9\% (TORGO). Harmful corrections remain low for high-confidence utterances (1.7\%, 5.0\%, 1.1\%), indicating effective confidence-guided behavior.

\begin{table}[h]
\centering
\caption{LLM correction performance by dataset and confidence level using dataset-specific mean thresholds. Help: correction improves transcription accuracy. Harm: correction reduces transcription accuracy.}

\label{tab:confidence-aware-corrections}
\resizebox{\linewidth}{!}{%
\begin{tabular}{llcccc}
\toprule
\textbf{Dataset} & \textbf{Conf} & \textbf{Avg Conf} & \textbf{Attempt (\%)} & \textbf{Help (\%)} & \textbf{Harm (\%)} \\
\midrule
\multirow{2}{*}{SAP-shared}   
 & \cellcolor{gray!10} Low  & \cellcolor{gray!10} 0.814 & \cellcolor{gray!10} 74.7 & \cellcolor{gray!10} 63.8 & \cellcolor{gray!10} 4.9 \\
 & High & 0.956 & 20.3 & 17.5 & 1.7 \\
\midrule
\multirow{2}{*}{SAP-unshared} 
 & \cellcolor{gray!10} Low  & \cellcolor{gray!10} 0.926 & \cellcolor{gray!10} 53.3 & \cellcolor{gray!10} 24.3 & \cellcolor{gray!10} 12.0 \\
 & High & 0.976 & 18.5 & 10.1 & 5.0 \\
\midrule
\multirow{2}{*}{TORGO}    
 & \cellcolor{gray!10} Low  & \cellcolor{gray!10} 0.954 & \cellcolor{gray!10} 30.9 & \cellcolor{gray!10} 9.6 & \cellcolor{gray!10} 5.1 \\
 & High & 0.989 & 4.9 & 2.4 & 1.1 \\
\bottomrule
\end{tabular}%
}
\end{table}

Qualitative examples in Fig.~\ref{fig:qualitative-examples} further illustrate this pattern and suggest potential underlying mechanisms of overcorrection. In the first case, the naive LLM replaces the correct "TEASED" with "ASKED," possibly reflecting frequency bias from training data where "ASK" lemma forms appear 424 times compared to only 7 TEASE lemma forms (61:1 ratio). The confidence-guided model preserves the original high-confidence word. In the second case, the naive LLM exhibits grammaticality bias, changing "WHAT" the more grammatically standard "WHAT'S". By contrast, the confidence-guided model respects the high-confidence word. These examples demonstrate how confidence guidance enables more selective corrections, preserving high-confidence words while allowing necessary semantic changes ("TRAINED" to "TRADING").

\begin{figure}[h]
\centering
\footnotesize
\begin{adjustbox}{max width=\linewidth}
\ttfamily
\begin{tabular}{lcccccc}
REF:        & CUB & BEAR & \textcolor{green!50!black}{TEASED} & HIS & PAPA \\
ASR:        & CUB & BEAR & \textcolor{green!50!black}{TEASED} & HIS & PAPA \\
Conf:       & [0.91] & [0.94] & [0.99] & [1.00] & [0.97] \\
Naive LLM:  & CUB & BEAR & \textcolor{red}{ASKED} & HIS & PAPA \\
CONF. LLM:& CUB & BEAR & \textcolor{green!50!black}{TEASED} & HIS & PAPA \\
\end{tabular}
\end{adjustbox}

\vspace{0.50em}

\begin{adjustbox}{max width=\linewidth}
\ttfamily
\begin{tabular}{lcccccc}
REF:        & \textcolor{green!50!black}{WHAT} & APPLE & \textcolor{green!50!black}{TRADING} & AT & \\
ASR:        & \textcolor{green!50!black}{WHAT} & APPLE & \textcolor{red}{TRAINED} & AT & \\
Conf:       & [0.99] & [0.93] & [0.96] & [0.94] & & \\
Naive LLM:  & \textcolor{red}{WHAT'S} & APPLE & \textcolor{green!50!black}{TRADING} & AT & \\
CONF. LLM:& \textcolor{green!50!black}{WHAT} & APPLE & \textcolor{green!50!black}{TRADING} & AT & \\
\end{tabular}
\end{adjustbox}

\caption{Qualitative examples showing that naive LLM correction overcorrects high-confidence words, while confidence-guided prompting avoids overcorrection.}
\label{fig:qualitative-examples}
\end{figure}


\section{Conclusion}

We introduced a confidence-guided framework for ASR error correction in impaired speech using large language models, where word-level uncertainty estimates from Tsallis entropy guide targeted corrections. This approach reduces overcorrection and improves reliability. 

Our results show that confidence information benefits correction with WER reductions on both spontaneous speech (SAP-unshared: 9.94\% to 9.47\%) and unseen datasets (TORGO: 10.83\% to 10.58\%). Our method generalizes across ASR architectures, achieving improvements on Whisper outputs (13.10\% to 4.19\% on SAP-shared) despite training only on Parakeet data, demonstrating robustness to different error patterns. Confidence-based prompting also improved correction balance by reducing harmful edits on high-confidence utterances while increasing helpful corrections on low-confidence ones. 

While promising, limitations include the SAP dataset's focus on mild-to-moderate PD speakers reading structured commands, which may limit generalizability to severe dysarthria and spontaneous speech. Additionally, entropy-based confidence measures require careful parameter tuning (alpha values, aggregation strategies) and may not be well-calibrated across different conditions.

\bibliographystyle{IEEEbib}
\bibliography{paper}

\end{document}